# A novel interpretable machine learning system to generate clinical risk scores: An application for predicting early mortality or unplanned readmission in a retrospective cohort study

Yilin Ning[1], Siqi Li[1], Marcus Eng Hock Ong[2,3,4], Feng Xie[1,2], Bibhas Chakraborty[1,2,5,6], Daniel Shu Wei Ting[1,7,8], Nan Liu[1,2,3,8,9*]

[1] Centre for Quantitative Medicine, Duke-NUS Medical School, Singapore, Singapore
[2] Programme in Health Services and Systems Research, Duke-NUS Medical School, Singapore, Singapore
[3] Health Services Research Centre, Singapore Health Services, Singapore, Singapore
[4] Department of Emergency Medicine, Singapore General Hospital, Singapore, Singapore
[5] Department of Statistics and Data Science, National University of Singapore, Singapore, Singapore
[6] Department of Biostatistics and Bioinformatics, Duke University, Durham, NC, United States
[7] Singapore Eye Research Institute, Singapore National Eye Centre, Singapore, Singapore
[8] SingHealth AI Health Program, Singapore Health Services, Singapore, Singapore
[9] Institute of Data Science, National University of Singapore, Singapore, Singapore

*Correspondence: Nan Liu, Programme in Health Services and Systems Research, Duke-NUS Medical School, 8 College Road, Singapore 169857, Singapore. Phone: +65 6601 6503. Email: liu.nan@duke-nus.edu.sg

**Abstract**

Risk scores are widely used for clinical decision making and commonly generated from logistic regression models. Machine-learning-based methods may work well for identifying important predictors, but such 'black box' variable selection limits interpretability, and variable importance evaluated from a single model can be biased. We propose a robust and interpretable variable selection approach using the recently developed Shapley variable importance cloud (ShapleyVIC) that accounts for variability across models. Our approach evaluates and visualizes overall variable contributions for in-depth inference and transparent variable selection, and filters out non-significant contributors to simplify model building steps. We derive an ensemble variable ranking from variable contributions, which is easily integrated with an automated and modularized risk score generator, AutoScore, for convenient implementation. In a study of early death or unplanned readmission, ShapleyVIC selected 6 of 41 candidate variables to create a well-performing model, which had similar performance to a 16-variable model from machine-learning-based ranking.

# INTRODUCTION

Scoring models are widely used in clinical settings for assessment of risks and guiding decision making.[1] For example, the LACE index[2] for predicting unplanned readmission and early death after hospital discharge has been applied and validated in various clinical and public health settings since it was developed in 2010.[3–5] The LACE index is appreciated for its simplicity, achieving moderate discrimination power[6] by using only four basic components: inpatient length of stay (LOS), acute admission, comorbidity, and the number of emergency department (ED) visits in past 6 months. There have been ongoing efforts to derive new readmission risk prediction models for improved performance and for specific subcohorts, which, despite the increasing availability of machine learning methods in recent years, are dominated by the traditional logistic regression approach.[7,8]

Logistic regression models generate transparent prediction models that are easily converted to risk scores, e.g., by rounding the regression coefficients,[9] where model interpretability is mainly reflected by variable selection. Traditional variable selection approaches (e.g., stepwise selection and penalized likelihood) are straightforward, but may not be sufficient in identifying important predictors and controlling model complexity.[10,11] Machine learning methods are well-performing alternatives, e.g., a study used decision trees and neural networks to heuristically develop a 11-variable logistic regression model from 76 candidate variables,[12] and a recently developed automated risk score generator, the AutoScore,[13] uses the random forest (RF) to rank variables and built well-performing scoring models using less than 10 variables in clinical applications.[13,14] Such machine learning methods rank variables by their importance to best-performing model(s) trained from data, but the 'black box' nature of most machine learning models limits the interpretability of the variable selection steps. The interpretability issue may be alleviated via post-hoc explanation, e.g., the widely used Shapley additive explanations (SHAP).[15] However, another general issue in current variable importance analyses has been largely ignored: restricting the assessment to best-performing models can lead to biased perception on variable importance.[16]

Since predictive performance is often not the only concern in practice, a recent study extended the investigation to a group of models that are 'good enough' to generate a variable importance cloud (VIC), which provides a comprehensive overview of how important (or unimportant) each variable could be to accurate prediction.[16] Our recent work combined VIC with the well-received Shapley values to devise a Shapley variable importance cloud (ShapleyVIC),[17] which easily integrates with current practice to provide less biased assessments of variable importance. In a recidivism risk prediction study, both ShapleyVIC and VIC found an overclaimed importance of race from single models,[16,17] but ShapleyVIC could easily compute uncertainty intervals for overall variable importance for statistical assessments. In a mortality risk prediction study using logistic regression,[17] ShapleyVIC handled strong collinearity among variables and effectively communicated findings through statistics and visualizations. These applications suggest ShapleyVIC as a suitable method for guiding variable selection steps when developing clinical scores from regression models.

In this paper, we explore the use of ShapleyVIC as an interpretable and robust variable selection approach for developing clinical scores. Specifically, we use an electronic health record dataset from ED to predict death or unplanned readmission within 30 days after hospital discharge. We interpret the importance of candidate variables with respect to a group of 'good' logistic regression models using visualizations of ShapleyVIC values, combine information across models to generate an ensemble variable ranking, and use the estimated overall importance to filter out candidate variables that will likely add noise to the scoring model. To develop scoring models from ShapleyVIC-ranked candidate variables, we take advantage of the disciplined and modularized AutoScore framework, which grows a relatively simple, interpretable model based on a ranked list of variables until the inclusion of any additional variable makes little improvement to predictive performance.[13] The current AutoScore framework ranks variables using the RF, which is a commonly used machine learning method in clinical applications that performs well and is easy to train, and its preferrable performance in identifying important variables over traditional statistical methods has been previously demonstrated.[13] Hence, in this work we include RF-based scoring models as a baseline to evaluate the ShapleyVIC-based model and to demonstrate the benefits of ShapleyVIC that are not available from existing variable importance methods. We also compare these scoring models to the LACE index to show improvements over existing models.

## RESULTS

**Study cohort**

This study aimed to develop a scoring model to predict unplanned readmission or death within 30 days after hospital discharge. The data was derived from a retrospective cohort study of cases who visited the ED of Singapore General Hospital and were subsequently admitted to the hospital. The full cohort consists of 411,137 cases, where 388,576 cases were eligible. 63,938 (16.5%) eligible cases developed the outcome of interest. As summarized in Table 1, cases with and without the outcome were significantly different (i.e., had p-value<0.05) in all 41 candidate variables except the use of ventilation. Specifically, compared to those without the event, cases with events tended to be older and have shorter ED LOS, higher ED triage, longer inpatient LOS, more ED visits in the past 6 months and more comorbid. The training, validation and test sets consisted of 272,004 (70%), 38,857 (10%) and 77,715 (20%) cases randomly drawn from the full cohort, respectively.

**RF-generated models**

We first describe scoring models generated using RF-based variable ranking, which ranks the 41 variables based on their importance to the RF trained on the training set and uses it in the variable ranking module of the AutoScore framework. Figure 1 presents the parsimony plot from AutoScore, which visualizes the improvement to model performance (measured using the area under the receiver operating characteristic curve [AUC] on the validation set) when each variable enters the scoring model. The AutoScore guideline suggested two feasible models: Model 1A using the first 6 variables (i.e., number of ED visits in past 6 months, inpatient LOS, ED LOS, ED boarding time, creatinine, and age) where the inclusion of the next variable only improved AUC by 0.2%, and Model 1B using the first 16 variables to include the 16$^{th}$ variable (i.e., metastatic cancer) that resulted in a pronounced increase (1.8%) in AUC. The six-variable Model 1A had an AUC of 0.739 (95% confidence interval [CI]: 0.734-0.743; see Table 2) evaluated on the validation set, comparable to that of the LACE index (AUC=0.733, 95% CI: 0.728-0.738), whereas Model 1B outperformed the LACE index with an AUC of 0.759 (95% CI: 0.754-0.764). Table 3 presents the scoring tables for Model 1A and 1B after fine-tuning the cut-off values for continuous variables, and the scoring table of the LACE index is provided as Appendix Table 1A for readers' convenience.

**ShapleyVIC-generated model**

Next, we describe the assessment of overall variable importance using ShapleyVIC values and the implications, the resulting ensemble ranking that accounts for variabilities in variable importance across models, the simplification of model building steps based on the significance of overall variable importance, and the final scoring model developed. Using the training set, we trained a logistic regression model on all 41 candidate predictors by minimizing model loss, and generated 350 nearly optimal models whose loss was no more than 5% higher than the minimum loss. We used the first 3500 cases in the validation set to evaluate the ShapleyVIC values of these 350 models, which we found enough for the algorithm to converge and generate stable values in preliminary experiments.

Unlike machine-learning-based methods (e.g., RF as described in the previous subsection) that focus on relative importance of candidate variables for ranking purpose, ShapleyVIC quantifies the extent of variable importance for more in-depth inference, and communicates the findings through different forms of visualizations to facilitate interpretation. Figure 2 visualizes the average ShapleyVIC values across 350 models and the 95% prediction intervals as estimates for overall variable importance and its variability, where non-positive values indicate unimportance and are visualized using grey bars. Figure 3 presents the distribution of ShapleyVIC values from individual models to visualize the relationship between variable importance and model performance among the nearly optimal models. While both the RF (see Figure 1) and the average ShapleyVIC values (see Figure 2) suggested the number of previous ED visits as the most important variable among the 41 candidates, the 95% PI of the average ShapleyVIC value (see Figure 2) further concluded the significance of its overall importance, and the violin plot in Figure 3 revealed its much higher importance than other variables in all nearly optimal models studied. Metastatic cancer had the second highest average ShapleyVIC value among all variables, consistent with its considerable contribution to model performance observed from the RF-based variable ranking.

Both metastatic cancer and age were important to all nearly optimal models studied (indicated by positive ShapleyVIC values), but the wider spread of ShapleyVIC values for age resulted in a wider 95% PI and hence a higher uncertainty for the average value. Similarly, although admission type had similar overall importance (indicated by the average ShapleyVIC value) as neighboring variables (cancer and systolic blood pressure), the wider spread and long left tail for admission type resulted in a 95% PI containing zero, indicating a non-significant overall importance for this variable. Among the 41 candidate variables, 20 variables had non-significant overall importance and were excluded from subsequent model building steps.

To rank variables using ShapleyVIC values, we ranked the 41 variables within each of the 350 nearly optimal model, averaged across models to generate an ensemble ranking for all variables and subsequently focused on the 21 variables with significant overall importance. Unlike the parsimony plot from RF-based variable ranking (see Figure 1), the parsimony plot based on the ensemble ranking (see Figure 4) increased smoothly until the 6$^{th}$ variable entered the model, and the increments in model performance became small afterwards. The parsimony plot suggested a feasible model, Model 2, also with 6 variables: number of ED visits in past 6 months, metastatic cancer, age, sodium, renal disease, and ED triage. Four of these six variables were in common with the final models built using RF, but now metastatic cancer was selected automatically without the need for manual inspection. As a sensitivity analysis, we also generated a parsimony plot using the ensemble ranking of all 41 variables, visualized in Supplementary Figure S1. Although some variables with non-significant overall importance ranked higher than those with significant overall importance (e.g., admission type ranked 14$^{th}$ among all 41 variables, higher than ED LOS and the next seven variables in Figure 4), these non-significant variables had small incremental change to model performance and did not affect the development of Model 2. Hence, the exclusion of variables with non-significant overall importance simplified model building steps without negative impact on the performance of the final model. The scoring table of Model 2 (after fine-tuning) is shown in Table 4, and the AUC evaluated on the test set (0.756, 95% CI: 0.751-0.760) was comparable to that of the 16-variable Model 1B and significantly higher than the LACE index (see Table 2).

## DISCUSSION

Identifying important variables to the outcome is crucial for developing well-performing and interpretable prediction models,[18] especially when developing clinical risk scores based on the simple structure of logistic regression models. The recently developed ShapleyVIC method[17] comprehensively assesses variable importance to accurate predictions, and quantifies the variability in importance to facilitate rigorous statistical inference. In this study, we propose ShapleyVIC as a variable selection tool for developing clinical risk scores, and illustrated its application in conjunction with the modularized AutoScore framework[13] by developing a risk score for unplanned readmission or death within 30 days after hospital discharge. Based completely on the scoring-generating logistic regression models, ShapleyVIC avoids bias in importance assessments due to the choice of variable ranking methods, and presents rich statistics on variable contributions through effective visualizations to support in-depth interpretation and guide variable selection. The ShapleyVIC-generated model used six of the 41 candidate variables, which outperformed the widely used LACE index and had comparable performance as a 16-variable model developed from RF-based variable ranking. Our work makes a novel contribution to the development of clinical risk scoring systems by providing an effective and robust variable selection method that is supported by statistical assessment and of variable importance and is tailored to score-generating regression models.

Models that are 'good enough' can be as relevant in real-life clinical applications as the 'best-performing' model, especially with respect to practical considerations such as clinical meaningfulness and costs.[16,17] The variable importance cloud proposed by Dong and Rudin[16] was the first to formally extend variable importance to such a group of models with nearly optimal performance. ShapleyVIC creates an ensemble of variable importance measures from a group of regression models using the state-of-the-art Shapley-based interpretable machine learning method, explicitly evaluates the variability of variable importance across models to support formal assessment of overall importance, and conveys such information through effective visualizations. In this work, we further propose a disciplined approach that takes into account such rich information on variable importance when ranking variables, and demonstrated its easy integration with the existing AutoScore framework[13] for automated development of scoring systems. In addition to using the average ShapleyVIC values across models as a measure of overall variable importance, we propose to use them as screening tools to exclude candidate variables that are less useful.

Although machine learning methods (e.g., RF and neural network) can be more efficient than traditional variable selection methods,[12,13,19] they often assume complex non-linear and/or tree-based structures that differ drastically from the linear structure assumed by scoring-generating logistic regression. Our application highlights such misalignment, where an important contributor (metastatic cancer) to the scoring model was only ranked 16[th] among all 41 variables by the RF, and the 4[th]-ranking variable (ED boarding time) by the RF had minimal incremental contribution to the performance of the scoring model. Such misalignment is less problematic to the AutoScore framework, as it can be observed from the parsimony plot and handled by manual adjustment to the variable list, e.g., by excluding ED boarding time from Model 1A and adding metastatic cancer to build a new 6-variable model with comparable performance (AUC=0.756, 95% CI: 0.751-0.760) to the 16-variable Model 1B. ShapleyVIC avoids such subjective assessments by providing an objective and data-driven ranking approach that is tailored to score-generating logistic regressions. Using an ensemble variable ranking across 350 well-performing logistic regression models, ShapleyVIC successfully assigned high ranks to all important contributors (including metastatic cancer), and excluded ED boarding time among other 20 variables that had non-significant overall importance. The model using the first six variables ranked by ShapleyVIC achieved a similar AUC to the 16-variable model based on RF.

Four of the six variables used in Model 2 (i.e., number of ED visits in past 6 months, metastatic cancer, age, and sodium) were also used in the 16-variable Model 1B. Both models outperformed the LACE index by using fewer variables, as the CCI used in LACE requires information on 17 comorbidities. A notable difference is the role of inpatient LOS, which is included in Model 1B and LACE index but not in Model 2. The two variables used in Model 2 but not the other models, i.e., renal disease and ED triage, were ranked 22$^{nd}$ and 23$^{rd}$ by RF, respectively. Creatinine, which was used in Model 1B but not Model 2, is reflective of renal function to some extent.[20] The use of ED triage in Model 2 is reasonable in our setting, where the full cohort consists of emergency admissions (i.e., inpatient admissions after ED visits), and ED triage is an important predictor for short-term clinical outcomes such as 30-day mortality or readmission.[21] However, scoring models discussed above were developed for illustrative purpose and should not be use in clinical applications without further investigations and validation.

Interestingly, when using the same variables, RF had a slightly lower AUC than Model 1B (AUC=0.754, 95% CI: 0.749-0.758) and a much lower AUC than Model 2 (AUC=0.663, 95% CI: 0.659-0.668). This again highlights the drastic difference between RF and logistic regression models and their perception on variable importance, and that complex and robust machine learning models do not necessarily outperform simple scoring models. However, compared to variable ranking based on a single RF, ShapleyVIC requires much longer run-time and larger memory space, which would benefit from the use of high-performance computers and parallel computing (option available in the ShapleyVIC R package[22]). Another limitation of ShapleyVIC is that the sampled set of models generated using our pragmatic sampling approach may not fully represent the entire set of near-optimal models (which is formally referred to as the Rashomon set),[17] which remains a challenge in this area of research.[1,16] In addition, although collinearity was not present in our example (where generalized VIF for all 41 candidate variables were below 2), it is generally an unresolved difficulty in variable importance assessments.[16,17,23–25] Our pragmatic solution of using absolute SAGE values as model reliance measures for variables with VIF>2 worked well in a previous empirical experiment,[17] and future work aims to devise more formal solutions.

Our work contributes to the recent emphasis on interpretability and transparency of prediction models for high-stakes decision making,[26] by devising a robust and interpretable approach for developing clinical scores. Unlike existing statistical or machine learning methods for variable selection, which focus on ranking variables by their importance to the prediction, ShapleyVIC rigorously evaluates variable contributions to the scoring model and visualizes the findings to enable a robust and fully transparent variable ranking procedure. The ShapleyVIC-based variable ranking, which is tailored to the score-generating logistic regression, is easily applied to the AutoScore framework for generating sparse scoring models with the flexibility to adjust for expert opinions and practical conventions. ShapleyVIC and AutoScore combines into an integrated approach for future interpretable machine learning research in clinical applications, which provides a disciplined solution to detailed assessment of variable importance and transparent development of clinical risk scores, and is easily implemented by executing a few commands from the respective R packages.[22,27] Although we have illustrated our approach for predicting early death or unplanned readmission, it is not limited to any specific clinical application. Moreover, the model-agnostic property of ShapleyVIC[17] and the recent extension of the AutoScore framework to survival outcomes[28] enables the use of our approach for a wider range of applications.

## METHODS

**AutoScore framework for developing risk scoring models**

The AutoScore framework provides an automated procedure for developing risk scoring models for binary outcomes. Interested readers may refer to the original paper[13] and the R package documentation[27] for detailed description of AutoScore methodology, and to a clinical application for an example use case.[14] In brief, AutoScore divides data into training, validation and test sets (typically consisting of 70%, 10% and 20% of total sample size, respectively), and ranks all candidate variables based on their importance to a RF trained on the training set. Continuous variables are then automatically categorized (using percentiles or k-means clustering)[13] for preferable clinical interpretation and to account for potential non-linear relationships between predictors and outcomes. AutoScore creates scoring models based on the logistic regression, where coefficients are scaled and rounded to non-negative integer values for convenient calculation. Following the RF-based variable ranking, AutoScore grows the scoring model by adding one variable each time and inspects the resulting improvement in model performance (measured using the AUC on the validation set) using a parsimony plot. The final model is often determined by selecting the top few variables where a reasonable AUC is reached and inclusion of an additional variable only results in a small increment (e.g., <1%) to AUC. After selecting the final list of variables, cut-off values used to categorize continuous variables may be fine-tuned to suit clinical practice and conventions, and the finalized model is evaluated on the test set.

**Shapley variable importance cloud (ShapleyVIC)**

In practical prediction tasks, researchers are often willing to select a simpler and interpretable model even when its performance is marginally lower than a more complex model. Hence, conventional variable importance assessments based on a single best model can be limiting. ShapleyVIC[17] is suitable for such purposes and assesses the overall contribution of variables by studying a group of models with near-optimal performance. In the case of developing scoring models for binary outcomes, near-optimal models can be characterized by vectors of regression coefficients corresponding to logistic loss no more than 5% higher than the minimum logistic loss.

ShapleyVIC consists of three general steps. First, a suitable number (e.g., 350) of near-optimal models are sampled from a multivariable normal centered by the regression coefficients of the optimal logistic regression model. The sampling steps are implemented in the ShapleyVIC package[22] and described in detail in the original paper.[17] Next, ShapleyVIC measures reliance of each nearly optimal model on variables using the Shapley additive global importance (SAGE) method, which measures variable impact on a model using Shapley values and is closely related to the commonly used SHAP method.[24] As explained in the original paper on ShapleyVIC, we use the absolute SAGE values instead of the unadjusted values to measure model reliance for variables involved in collinearity, identified by variable inflation factor above two from the optimal logistic regression model. Finally, the ShapleyVIC values are pooled across all nearly optimal models using a random-effect meta-analysis approach, which quantifies the overall importance of variables and explicitly evaluates the variability of these measures across well-performing models (in terms of a 95% PI for a new model) to support formal statistical inference. The average ShapleyVIC values and 95% PIs are visualized using a bar plot, and the variability of ShapleyVIC values across models is highlighted using a colored violin plot for additional insights.

**Ensemble variable ranking from ShapleyVIC**

To apply ShapleyVIC in practical development of scoring models based on logistic regression models, we use an ensemble variable ranking approach from ShapleyVIC that is tailored to score-generating models and filter out variables that are not likely to make significant contributions. First, we describe our proposed ensemble variable ranking for $d$ variables for a single nearly optimal model $f$. Let $\widehat{mr}_j^s(f)$ denote the ShapleyVIC value of the $j$-th variable ($j = 1, \dots, d$) for model $f$, which, as described above, is either the SAGE value or the absolute SAGE value. The variability of $\widehat{mr}_j^s(f)$ can therefore be measured using the standard error of the SAGE value, denoted by $\sigma_j(f)$, that is readily available from the SAGE algorithm. Assuming independent normal distributions for SAGE values,[17,24] e.g., $X_j$ and $X_k$, to model $f$, based on the normal distribution of the difference between their ShapleyVIC values: $\widehat{mr}_j^s(f) - \widehat{mr}_k^s(f) \sim N\left(mr_j^s(f) - mr_k^s(f), \sigma_j^2(f) + \sigma_k^2(f)\right)$. We compare all possible pairs of variables, and subsequently rank the $d$ variables to model $f$ based on the number of times each variable has significantly larger ShapleyVIC value than the other $d-1$ variables. Assigning rank 1 to the most important variable and using the same smallest integer value available for tied ranks, we rank the $d$ variables for each model, and use the average rank of each variable across all nearly optimal models to generate an ensemble ranking.

Instead of considering all $d$ candidate variables in subsequent model building steps, we propose to filter out variables that are not likely important contributors to accurate predictions based on their overall importance, corresponding to variables where the 95% PI for the average ShapleyVIC value contains or is entirely below zero. In this work, we applied the proposed ensemble variable ranking to the AutoScore framework, using it to replace RF when ranking variables.

**Study design and data preparation**

Our empirical experiment was based on a retrospective cohort study of patients who visited the ED of Singapore General Hospital in years 2009 to 2017 and were subsequently admitted to the hospital. The outcome of interest was readmission or death within 30 days after discharge from hospital. Since data was collected from ED, all readmissions in the cohort were unplanned readmissions. Information on patient demographics, ED administration, inpatient admission, healthcare utilization, clinical tests, vital signs, and comorbidities was extracted from the hospital electronic health record system. A full list of variables is provided in Table 1. ED triage was evaluated using the Patient Acuity Category Scale,[21] the current triage system used in Singapore. Admission type refers to the four types of wards in the hospital with different costs,[29] which partially reflects the socio-economic status of patients. We excluded patients who were not Singapore residents, died in hospital or aged below 21. The final cohort was split into training (70%), validation (10%) and testing sets (20%), and missing values for vital signs or clinical tests were imputed using the median value in the validation set. To compute the LACE index, we computed the CCI as described in a previous work.[30]

## DATA AVAILABILITY

Data are not available because they contain sensitive patient information from electronic health records. De-identified data is available from the corresponding author upon reasonable request.

## CODE AVAILABILITY

Source code is available from https://github.com/nliulab/ShapleyVIC.

**ACKNOWLEDGEMENT**

Yilin Ning is supported by the Khoo Postdoctoral Fellowship Award (Project No. Duke-NUS-KPFA/2021/0051) from the Estate of Tan Sri Khoo Teck Puat.

**AUTHOR CONTRIBUTIONS**
NL and YN conceptualized the study and designed the analytical method. YN developed the software package for the method, conducted data analyses and drafted the manuscript. SL participated in the development of software package and data analyses. All authors interpreted the data, discussed the results, and critically revised the manuscript for intellectual content. NL supervised the study.

**COMPETING INTERESTS**
The authors declare no competing interests.

**Table 1. Descriptive statistics of the full cohort. The outcome is death or readmission within 30 days after hospital discharge.**

|  | Overall (N=388,576) | With outcome (N=63,938, 16.5%) | Without outcome (N=324,638, 83.5%) | p-value** |
|---|---|---|---|---|
| **Patient demographics** | | | | |
| Age (years): mean (SD) | 62.0 (17.8) | 67.6 (15.3) | 60.9 (18.0) | <0.001 |
| Gender: n (%) | | | | <0.001 |
|   Female | 195426 (50.3) | 30405 (47.6) | 165021 (50.8) | |
|   Male | 193150 (49.7) | 33533 (52.4) | 159617 (49.2) | |
| Race: n (%) | | | | <0.001 |
|   Chinese | 274929 (70.8) | 48079 (75.2) | 226850 (69.9) | |
|   Indian | 41718 (10.7) | 6277 (9.8) | 35441 (10.9) | |
|   Malay | 47528 (12.2) | 7430 (11.6) | 40098 (12.4) | |
|   Others | 24401 (6.3) | 2152 (3.4) | 22249 (6.9) | |
| **ED admission** | | | | |
| ED LOS (hours): mean (SD) | 2.8 (1.7) | 2.5 (1.6) | 2.8 (1.7) | <0.001 |
| ED triage: n (%) | | | | <0.001 |
|   P1 | 69383 (17.9) | 14751 (23.1) | 54632 (16.8) | |
|   P2 | 219245 (56.4) | 39287 (61.4) | 179958 (55.4) | |
|   P3 and P4 | 99948 (25.7) | 9900 (15.5) | 90048 (27.7) | |
| ED boarding time (hours): mean (SD) | 4.8 (3.7) | 4.8 (3.9) | 4.8 (3.7) | 0.014 |
| Consultation waiting time (hours): mean (SD) | 0.8 (0.8) | 0.7 (0.7) | 0.8 (0.8) | <0.001 |
| Day of week: n (%) | | | | <0.001 |
|   Friday | 54461 (14.0) | 8969 (14.0) | 45492 (14.0) | |
|   Monday | 64914 (16.7) | 10549 (16.5) | 54365 (16.7) | |
|   Weekend | 99797 (25.7) | 16903 (26.4) | 82894 (25.5) | |
|   Midweek | 169404 (43.6) | 27517 (43.0) | 141887 (43.7) | |
| **Inpatient admission** | | | | |
| Inpatient LOS (days): mean (SD) | 6.54 (11.66) | 8.19 (12.17) | 6.21 (11.52) | <0.001 |
| Admission type | | | | <0.001 |
|   A1 | 14964 (3.9) | 1401 (2.2) | 13563 (4.2) | |
|   B1 | 32688 (8.4) | 3144 (4.9) | 29544 (9.1) | |
|   B2 | 185244 (47.7) | 27713 (43.3) | 157531 (48.5) | |
|   C | 155680 (40.1) | 31680 (49.5) | 124000 (38.2) | |
| **Healthcare utilization: mean (SD)** | | | | |
| No. ED visits in past 6 months | 0.59 (1.42) | 1.58 (2.59) | 0.39 (0.93) | <0.001 |
| No. surgery in past year | 0.20 (0.75) | 0.43 (1.10) | 0.16 (0.65) | <0.001 |
| No. ICU stays in past year | 0.02 (0.26) | 0.05 (0.36) | 0.02 (0.23) | <0.001 |
| No. HD stays in past year | 0.09 (0.47) | 0.17 (0.69) | 0.07 (0.40) | <0.001 |
| **Vital sign and clinical tests at ED** | | | | |
| Ventilation: n (%) | 53 (0.0) | 8 (0.0) | 45 (0.0) | 0.789 |
| Resuscitation: n (%) | 4136 (1.1) | 874 (1.4) | 3262 (1.0) | <0.001 |
| Pulse, beat/minute: mean (SD)* | 82.33 (16.81) | 84.56 (17.51) | 81.89 (16.63) | <0.001 |
| Respiration, breath/minute: mean (SD)* | 17.86 (1.65) | 18.08 (1.90) | 17.81 (1.59) | <0.001 |
| $SpO_2$, %: mean (SD)* | 98.01 (3.05) | 97.87 (3.28) | 98.04 (3.00) | <0.001 |
| DBP, mmHg: mean (SD)* | 71.56 (13.36) | 70.60 (13.66) | 71.75 (13.29) | <0.001 |
| SBP, mmHg: mean (SD)* | 134.46 (25.32) | 134.23 (26.19) | 134.50 (25.14) | 0.013 |
| Bicarbonate, mmol/L: mean (SD)* | 22.80 (3.65) | 22.50 (4.16) | 22.86 (3.54) | <0.001 |
| Creatinine, μmol/L: mean (SD)* | 153.28 (207.43) | 203.71 (249.54) | 143.11 (196.30) | <0.001 |
| Potassium, mmol/L: mean (SD)* | 4.16 (0.71) | 4.25 (0.78) | 4.14 (0.69) | <0.001 |
| Sodium, mmol/L: mean (SD)* | 135.07 (5.02) | 134.13 (5.77) | 135.26 (4.83) | <0.001 |
| **Comorbidity: n (%)** | | | | |
| Myocardial infarction | 22064 (5.7) | 7408 (11.6) | 14656 (4.5) | <0.001 |
| Congestive heart failure | 43758 (11.3) | 13357 (20.9) | 30401 (9.4) | <0.001 |
| Peripheral vascular disease | 22239 (5.7) | 6883 (10.8) | 15356 (4.7) | <0.001 |
| Stroke | 50400 (13) | 11719 (18.3) | 38681 (11.9) | <0.001 |
| Dementia | 11178 (2.9) | 3291 (5.1) | 7887 (2.4) | <0.001 |
| Pulmonary | 37652 (9.7) | 9722 (15.2) | 27930 (8.6) | <0.001 |
| Rheumatic | 5507 (1.4) | 1213 (1.9) | 4294 (1.3) | <0.001 |
| Peptic ulcer disease | 14678 (3.8) | 3888 (6.1) | 10790 (3.3) | <0.001 |
| Mild liver disease | 18402 (4.7) | 4857 (7.6) | 13545 (4.2) | <0.001 |
| Severe liver disease | 5893 (1.5) | 2296 (3.6) | 3597 (1.1) | <0.001 |
| Diabetes (without complications) | 49057 (12.6) | 9918 (15.5) | 39139 (12.1) | <0.001 |
| Diabetes with complications | 96334 (24.8) | 22601 (35.3) | 73733 (22.7) | <0.001 |
| Paralysis | 21563 (5.5) | 5248 (8.2) | 16315 (5) | <0.001 |
| Renal | 80895 (20.8) | 22566 (35.3) | 58329 (18) | <0.001 |
| Cancer (non-metastatic) | 34385 (8.8) | 8532 (13.3) | 25853 (8) | <0.001 |
| Metastatic cancer | 28630 (7.4) | 11049 (17.3) | 17581 (5.4) | <0.001 |
| CCI: mean (SD) | 2.41 (2.68) | 4.14 (3.00) | 2.07 (2.48) | <0.001 |

*: Excluding 7521 missing entries for pulse, 8605 missing entries for respiration, 8533 missing entries for $SpO_2$, 4175 missing entries for DBP, 4181 missing entries for and SBP, 48739 missing entries for bicarbonate, 48695 missing entries for creatinine, 50358 missing entries for potassium, and 48637 missing entries for sodium.
**: P-values were reported from two-sample t-tests for continuous variables and Chi-squared tests for categorical variables.
CCI: Charlson comorbidity index; DBP: diastolic blood pressure; ED: emergency department; HD: high dependency ward; ICU: intensive care unit; LOS: length of stay; SBP: systolic blood pressure; SD: standard deviation; $SpO_2$: blood oxygen saturation.

**Table 2. Model performance evaluated on the test set.**

| Method | Model | Number of variables | AUC (95% CI) |
|---|---|---|---|
| LACE | -- | 3+17* | 0.733 (0.728, 0.738) |
| AutoScore | Model 1A | 6 | 0.739 (0.734, 0.743) |
| | Model 1B | 16 | 0.759 (0.754, 0.764) |
| AutoScore-ShapleyVIC | Model 2 | 6 | 0.756 (0.751, 0.760) |

*: LACE is computed from inpatient length of stay, number of ED visits in past 6 months, acute admission and Charlson comorbidity index (which is computed from 17 comorbidities). See Appendix Table A1 for detailed scoring table.

**Table 3. Scoring tables of Model 1A and 1B generated using AutoScore, with variables ranked by random forest.**

| Variable | Interval | Point Model 1A | Model 1B |
|---|---|---|---|
| Number of ED visits in past 6 months | <1 | 0 | 0 |
| | [1,3) | 17 | 11 |
| | >=3 | 40 | 27 |
| Inpatient LOS (days) | <1 | 0 | 0 |
| | [1,2) | 1 | 1 |
| | [2,7) | 6 | 3 |
| | >=7 | 12 | 6 |
| ED LOS | <40min | 12 | 5 |
| | [40min, 80min) | 9 | 3 |
| | [80min, 4h) | 6 | 2 |
| | [4h, 6h) | 2 | 1 |
| | >=6h | 0 | 0 |
| ED boarding time | <80min | 0 | 0 |
| | [80min, 6.5h) | 2 | 1 |
| | >=6.5h | 3 | 1 |
| Creatinine (μmol/L) | <45 | 7 | 2 |
| | [45, 60) | 1 | 0 |
| | [60, 135) | 0 | 1 |
| | [135, 600) | 7 | 5 |
| | >=600 | 8 | 7 |
| Age (years) | <25 | 0 | 0 |
| | [25,45) | 7 | 4 |
| | [45,75) | 18 | 10 |
| | [75,85) | 21 | 13 |
| | >=85 | 25 | 16 |
| Systolic blood pressure (mmHg) | <100 | -- | 3 |
| | [100, 115) | -- | 2 |
| | [115, 155) | -- | 1 |
| | >=155 | -- | 0 |
| Bicarbonate (mmol/L) | <17 | -- | 1 |
| | [17, 20) | -- | 1 |
| | [20, 28) | -- | 0 |
| | >=28 | -- | 3 |
| Pulse | <60 | -- | 0 |
| | [60, 70) | -- | 1 |
| | [70, 100) | -- | 2 |
| | >=100 | -- | 3 |
| Consultation waiting time (hours) | <1.5 | -- | 2 |
| | [1.5, 2.5) | -- | 1 |
| | >=2.5 | -- | 0 |
| Diastolic blood pressure (mmHg) | <50 | -- | 1 |
| | [50, 95) | -- | 0 |
| | >=95 | -- | 1 |
| Potassium (mmol/L) | <3.5 | -- | 2 |
| | [3.5, 4) | -- | 0 |
| | [4, 4.5) | -- | 1 |
| | >=4.5 | -- | 2 |
| Sodium (mmol/L) | <125 | -- | 6 |
| | [125,130) | -- | 4 |
| | [130,135) | -- | 2 |
| | >=135 | -- | 0 |
| $SpO_2$ | <95 | -- | 1 |
| | >=95 | -- | 0 |
| Respiration | <17 | -- | 0 |
| | [17, 20) | -- | 1 |
| | >=20 | -- | 3 |
| Metastatic cancer | No | -- | 0 |
| | Yes | -- | 15 |

"[A, B)" indicates an interval inclusive of the lower limit and exclusive of the upper limit.
"--" indicates variables not included in a model.
h: hours; min: minutes; ED: Emergency department; LOS: length of hospital stay.

**Table 4. Scoring tables of Model 2 generated using AutoScore, with variables ranked by ShapleyVIC.**

| Variable | Interval | Point Model 2 |
|---|---|---|
| Number of ED visits in past 6 months | <1 | 0 |
| | [1, 3) | 14 |
| | >=3 | 33 |
| Metastatic cancer | No | 0 |
| | Yes | 21 |
| Age (years) | <25 | 0 |
| | [25, 45) | 4 |
| | [45, 75) | 12 |
| | [75, 85) | 15 |
| | >=85 | 19 |
| Sodium (mmol/L) | <125 | 10 |
| | [125, 130) | 7 |
| | [130, 135) | 4 |
| | >=135 | 0 |
| Renal disease | No | 0 |
| | Yes | 8 |
| ED triage | P1 | 9 |
| | P2 | 6 |
| | P3 and P4 | 0 |

"[A, B)" indicates an interval inclusive of the lower limit and exclusive of the upper limit.
"--" indicates variables not included in a model.
h: hours; min: minutes; ED: Emergency department; LOS: length of hospital stay.

**Figure 1. Parsimony plot on validation set using random-forest-based variable ranking.**

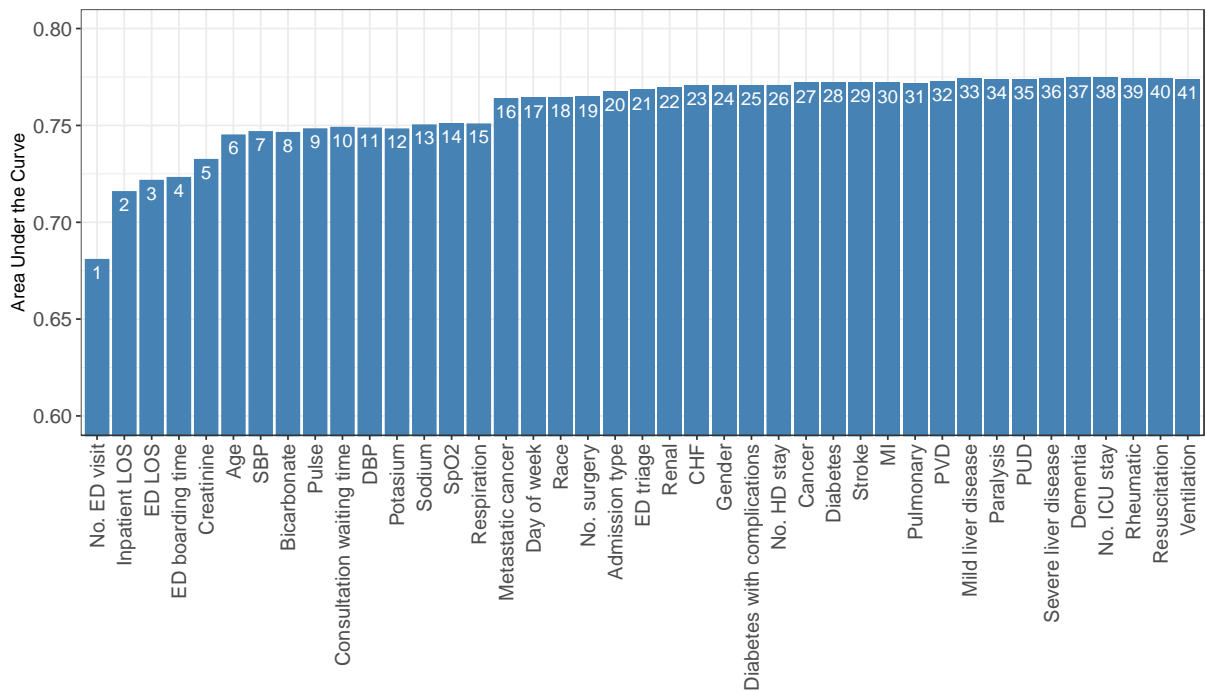

Number of ED visit is within 6 months before current inpatient stay. Number of surgery, ICU stay and HD stay are within 1 year before current inpatient stay.
CHF: Congestive heart failure; DBP: diastolic blood pressure; ED: emergency department; HD: high dependency ward; ICU: intensive care unit; LOS: length of stay; MI: Myocardial infarction; PVD: Peripheral vascular disease; PUD: Peptic ulcer disease; SBP: systolic blood pressure; SpO$_2$: blood oxygen saturation.

**Figure 2. Average ShapleyVIC values and 95% prediction intervals from 350 nearly optimal logistic regression models.**

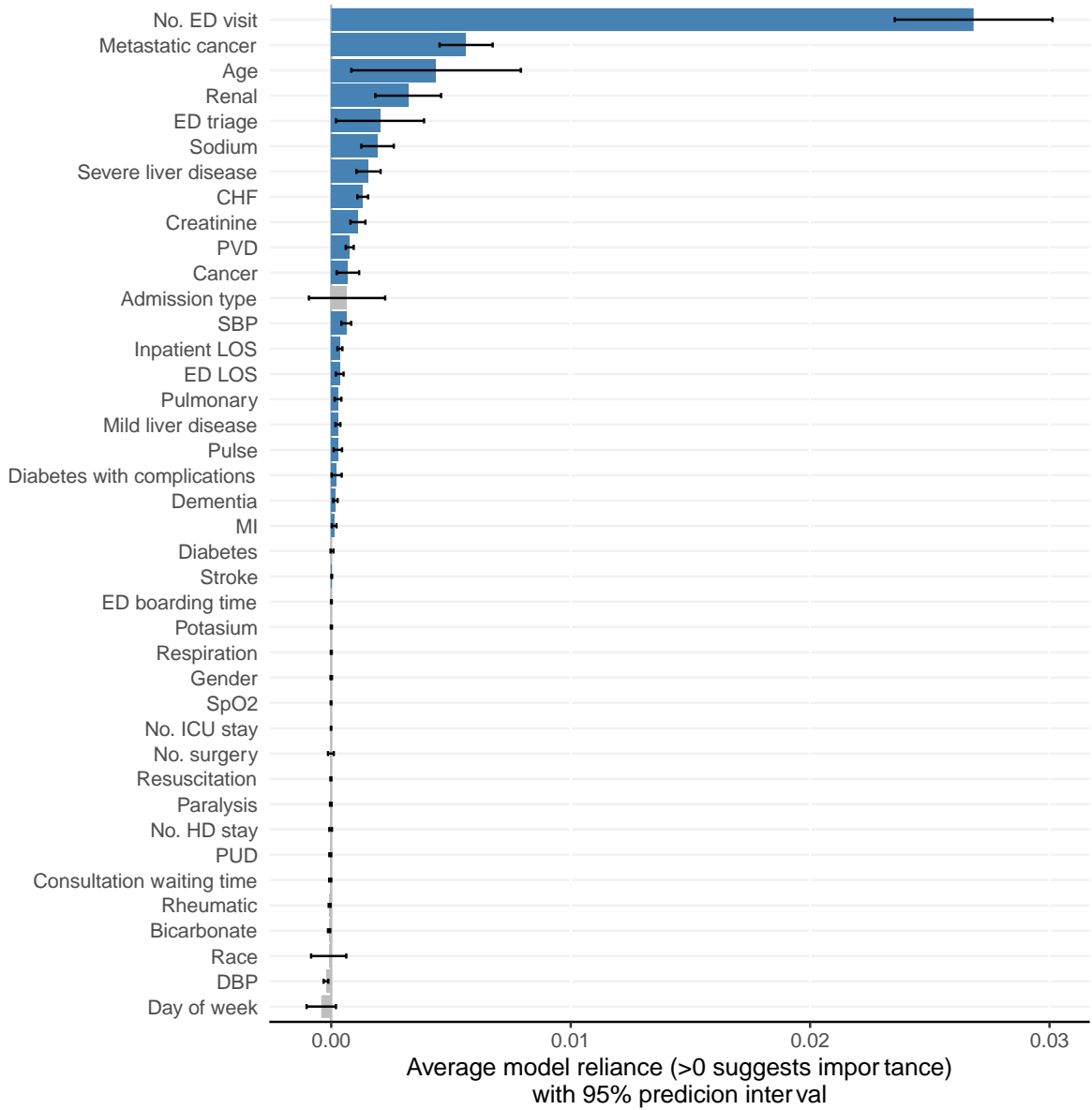

Number of ED visit is within 6 months before current inpatient stay. Number of surgery, ICU stay and HD stay are within 1 year before current inpatient stay.

CHF: Congestive heart failure; DBP: diastolic blood pressure; ED: emergency department; HD: high dependency ward; ICU: intensive care unit; LOS: length of stay; MI: Myocardial infarction; PVD: Peripheral vascular disease; PUD: Peptic ulcer disease; SBP: systolic blood pressure; SpO$_2$: blood oxygen saturation.

**Figure 3. ShapleyVIC values from 350 nearly optimal logistic regression models, arranged in descending order of average ShapleyVIC values.**

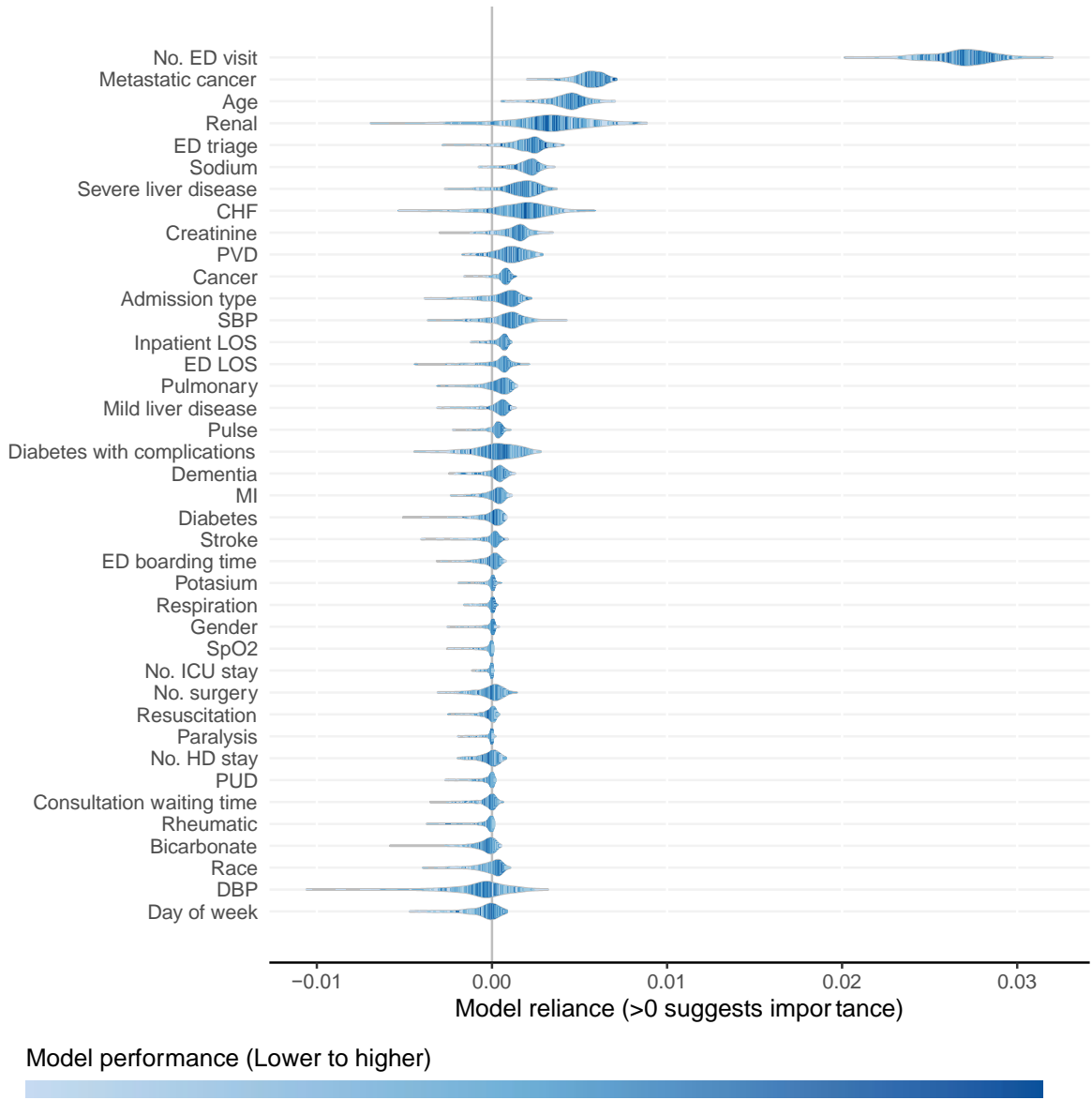

Number of ED visit is within 6 months before current inpatient stay. Number of surgery, ICU stay and HD stay are within 1 year before current inpatient stay.

CHF: Congestive heart failure; DBP: diastolic blood pressure; ED: emergency department; HD: high dependency ward; ICU: intensive care unit; LOS: length of stay; MI: Myocardial infarction; PVD: Peripheral vascular disease; PUD: Peptic ulcer disease; SBP: systolic blood pressure; SpO$_2$: blood oxygen saturation.

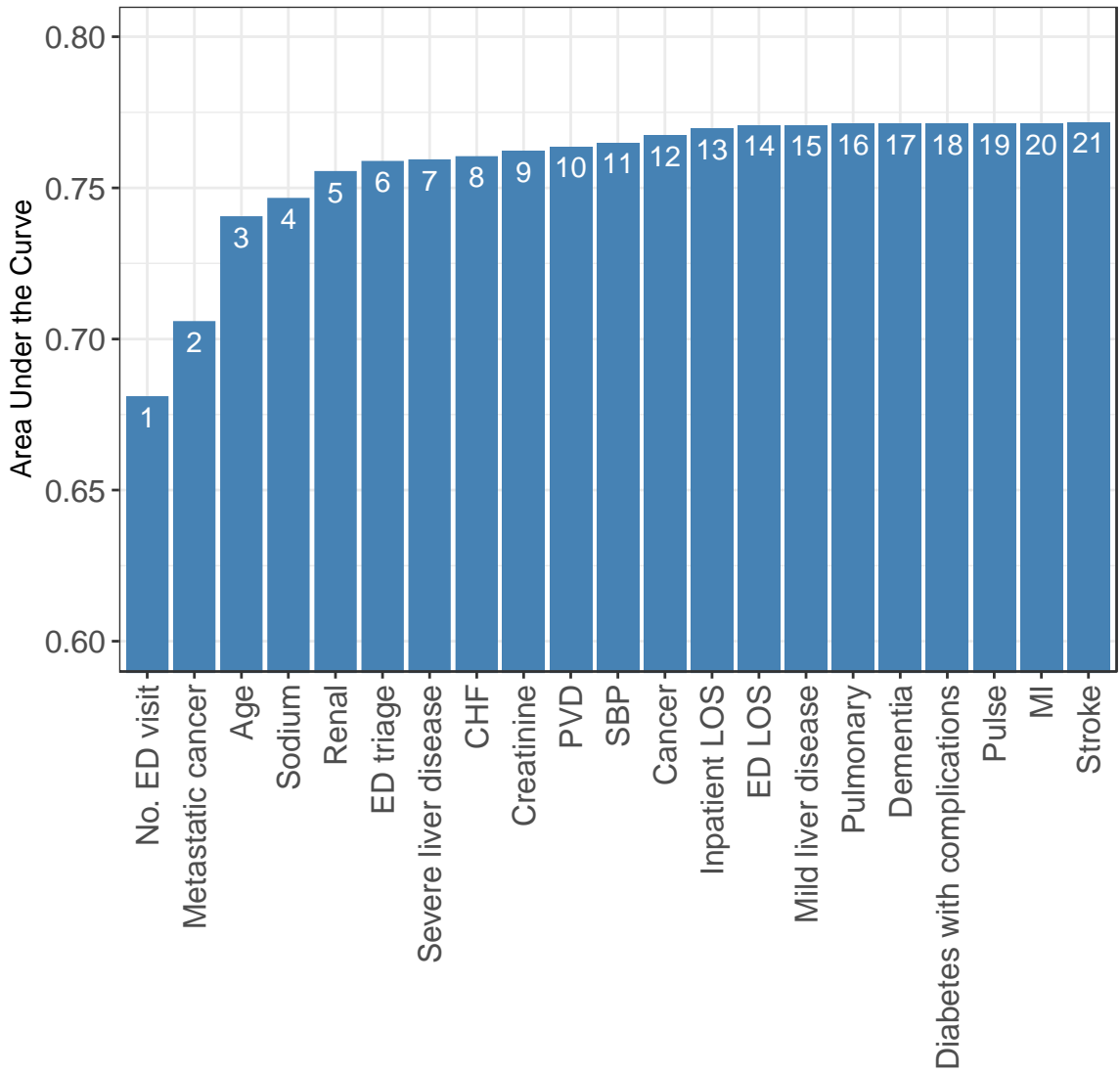

**Figure 4.** Parsimony plot on the validation set using ShapleyVIC variable ranking, based on 21 variables with significant average ShapleyVIC values. Note that important contributors to model performance, e.g., age and metastatic cancer, are ranked higher by ShapleyVIC than compared to the random-forest-based ranking in Figure 1.

**Supplementary information**

**Table S1. The LACE index.**

| Variable | Interval | Point |
|---|---|---|
| Inpatient length of stay | <1 | 0 |
| | 1 | 1 |
| | 2 | 2 |
| | 3 | 3 |
| | 4-6 | 4 |
| | 7-13 | 5 |
| | >=14 | 7 |
| Acute (emergent) admission | Yes | 3 |
| Charlson comorbidity index | 0 | 0 |
| | 1 | 1 |
| | 2 | 2 |
| | 3 | 3 |
| | >=4 | 5 |
| Visits to emergency department during previous 6 months | 0 | 0 |
| | 1 | 1 |
| | 2 | 2 |
| | 3 | 3 |
| | >=4 | 4 |

**Figure S1. Parsimony plot on the validation set using ShapleyVIC variable ranking, based on all 41 variables.**

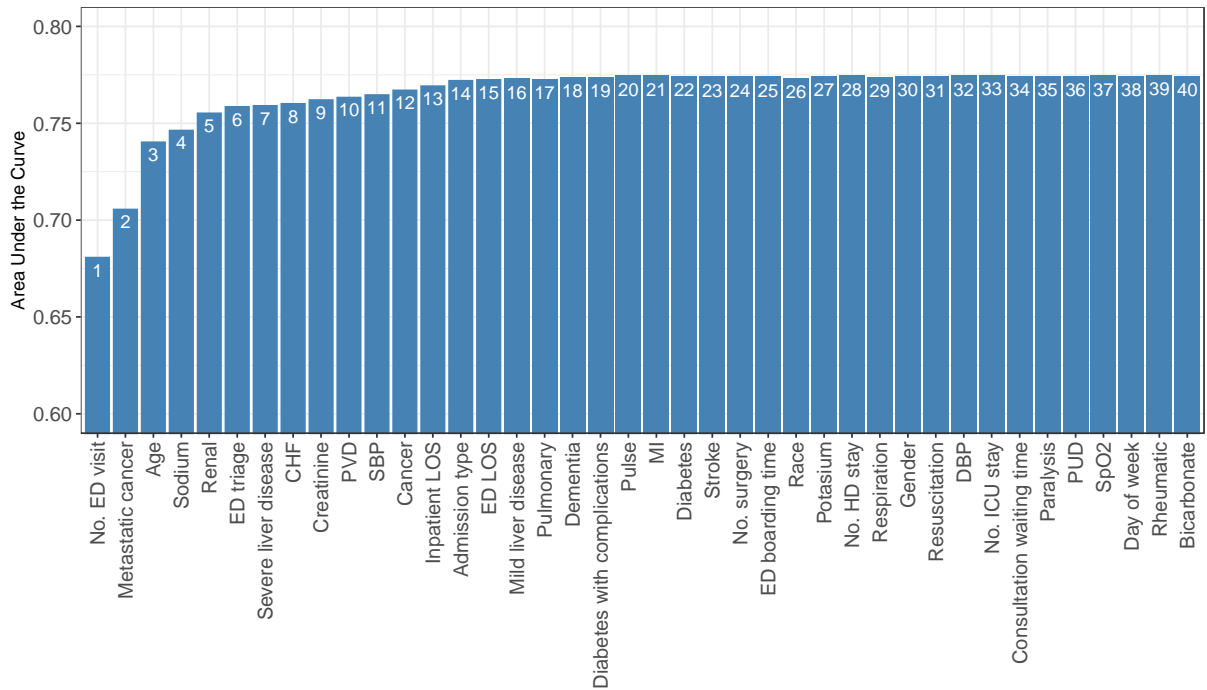